\documentclass{article}

\usepackage[preprint]{neurips_2023}

\usepackage[utf8]{inputenc} 
\usepackage[T1]{fontenc}    
\usepackage[hidelinks]{hyperref}       
\usepackage{url}            
\usepackage{booktabs}       
\usepackage{amsfonts}       
\usepackage{nicefrac}       
\usepackage{microtype}      
\usepackage{xcolor}         
\usepackage{graphicx}
\usepackage{subcaption}
\usepackage{todonotes}
\usepackage[inline]{enumitem}
\usepackage{xcolor}
\usepackage{amsthm}
\usepackage{multirow}
\usepackage{array}
\usepackage {tikz}
\usepackage{pifont}

\newlist{myenumerate}{enumerate}{1}
\setlist[myenumerate]{label=\textcolor{blue}{\textbf{\arabic*.}}}

\title{Understanding the Capabilities of Large Language Models for Automated Planning}

\author{%
  Vishal Pallagani \\
  AIISC, University of South Carolina \\
  \texttt{vishalp@mailbox.sc.edu} \\
  \And
  Bharath Muppasani \\
  AIISC, University of South Carolina \\
  \texttt{bharath@email.sc.edu} \\
  \And
  Keerthiram Murugesan \\
  IBM T.J. Watson Research Center \\
  \texttt{keerthiram.murugesan@ibm.com} \\
  \And
  Francesca Rossi \\
  IBM T.J. Watson Research Center \\
  \texttt{francesca.rossi2@ibm.com} \\
  \And
  Biplav Srivastava \\
  AIISC, University of South Carolina \\
  \texttt{biplav.s@sc.edu} \\
  \And
  Lior Horesh \\
  IBM T.J. Watson Research Center \\
  \texttt{lhoresh@us.ibm.com} \\
  \And
  Francesco Fabiano \\
  University of Udine \\
  \texttt{francesco.fabiano@unipr.it} \\
  \And
  Andrea Loreggia \\
  University of Brescia \\
  \texttt{andrea.loreggia@gmail.com} \\
}

\begin{document}

\maketitle

\begin{abstract}

Automated planning is concerned with developing efficient algorithms to generate plans or sequences of actions to achieve a specific goal in a given environment. Emerging Large Language Models (LLMs) can answer questions, write high-quality programming code, and predict protein folding, showcasing their versatility in solving various tasks beyond language-based problems. 
In this paper, we aim to explore how LLMs can also be used for automated planning. 
To do so, we seek to answer four key questions. Firstly, we want to understand the extent to which LLMs can be used for plan generation. Secondly, we aim to identify which pre-training data is most effective in facilitating plan generation. Thirdly, we investigate whether fine-tuning or prompting is a more effective approach for plan generation. Finally, we explore whether LLMs are capable of plan generalization. By answering these questions, the study seeks to shed light on the capabilities of LLMs in solving complex planning problems and provide insights into the most effective approaches for using LLMs in this context.

\end{abstract}

\section{Introduction}

Automated Planning \citep{book-planning-GhallabNauTraverso04} focuses on generating sequences of actions, called plans, for an agent to navigate from an initial state to a desired goal state. We represent the planning problems using the Planning Domain Definition Language (PDDL)~\citep{aeronautiques1998pddl}, a Lisp-inspired declarative language with the specifications for the (partial) transition model, initial \& goal states and constraints (such as preconditions and effects). Automated planners must follow these specifications to generate optimal or near-optimal plans, without violating any constraints. Recently, LLMs such as GPT-4 \citep{openai2023gpt4}, have demonstrated the ability to generate a executable piece of code in any programming language \citep{poldrack2023aiassisted}. Both PDDL and programming languages use a similar type of formal syntax and share common concepts such as variables, functions, and control structures. 
This notion of resemblance motivates us to investigate the capabilities of LLMs in plan generation. 

LLMs are built using neural networks with millions/billions of learnable parameters, pretrained with a large corpus of natural language data. LLMs such as GPT4 have been shown to generate human-like, coherent, and diverse texts. There has been recent interest in exploring the capabilities of LLMs beyond the applications in many natural language processing tasks \citep{llm-overview, zhao2023survey}. For example, in code generation \citep{wang2021codet5, feng2020codebert, chen2021evaluating}, protein folding \citep{unsal2022learning, ferruz2022controllable}, etc. However, one task that remains elusive for LLMs is automated planning \citep{valmeekam2022large}, as it requires reasoning about the effects of actions and finding optimal or near-optimal sequences of actions to achieve a desired goal.  Automated planning is crucial for many real-world applications, such as robotics, dialog systems, game playing, and more. Therefore, it is important to understand if and how the recent progress in LLMs can be leveraged for planning, including their limitations and opportunities for this task. 

In this paper, we provide a comprehensive analysis of LLMs' capabilities for automated planning. Toward this goal, we address the following four research questions: 
\begin{enumerate*}[label=\textbf{(\arabic*)}]
    \item To what extent can LLMs solve planning problems?
    \item What pre-training data is effective for plan generation?
    \item Does fine-tuning and prompting improve LLM's plan generation?
    \item Are LLMs capable of plan generalization? 
    \end{enumerate*}
To answer these questions, we compare different LLMs on six classical planning domains using both fine-tuning and prompting approaches.
We propose a metric to measure plan generalization and also introduce three new tasks to evaluate LLMs on plan generalization. Despite the inapt claims on LLMs for automated planning \citep{valmeekam2022large}, we show favorable outcomes with appropriate selection of LLM, data preparation, and fine-tuning.  We claim that LLMs pre-trained on code generation can benefit from further fine-tuning with problems from several automated planning domains, although their generalization capabilities seem limited. We recommend further research in LLM for better plan generalization.
 
Our key contributions in this paper are:
 \begin{enumerate*}[label=\textbf{(\alph*)}]
\item a diverse set of benchmark problems to evaluate LLMs for automated planning (along with a publicly available codebase with model weights to drive future research).
\item a metric to evaluate plan generalization and design new tasks to measure it.
\item a thorough empirical analysis of LLMs on planning-related metrics and insights on an appropriate selection of LLMs for automated planning.
\end{enumerate*}

\section{Background and Related Work}


This section reports a brief review of the pertinent literature regarding exploring the use LLMs for planning. In Table \ref{terminology}, we summarize the planning-related terminology for a better understanding.

\begin{table}
  \caption{Definitions for the terms used in the paper.}
  \scriptsize
  \label{terminology}
  \centering
  \resizebox{\textwidth}{!}{%
  \begin{tabular}{p{4cm}p{9cm}}
    \toprule
    Term & Definition\\
    \midrule
    \textbf{Classical planning problem} & It is a 4-tuple $\langle S, s_0, A, G\rangle$ where $S$ represents the set of all possible states of the planning problem, $s_0$ is the initial state, $A$ is the set of all possible actions that can be performed in the planning problem, and $G$ represents the set of all goal states that the agent is trying to reach.\\
    \midrule
    \textbf{PDDL} & A formal language used to describe classical planning problems. It requires a domain and problem file.\\
    \midrule
    \textbf{Domain File} & Defines the set of actions, along with their preconditions and effects, objects and their relations, and predicates that can be used to describe a planning problem within a specific domain.\\
    \midrule
    \textbf{Problem File} & Define the initial state of a planning problem, along with the goal state(s) that needs to be achieved.\\
    \midrule
    \textbf{Planner} & An algorithmic tool that generates a plan of actions to achieve a desired goal state, given the domain and problem PDDL files. An example is the tool FastDownward~\citep{helmert2006fast}.\\
    \midrule
    \textbf{Plan} & A sequence of actions that transforms the initial state into one that respects the goal conditions. \\
    \midrule 
    \textbf{Satisficing plan} & A plan that achieves the goal state.\\
    \midrule 
    \textbf{Optimal plan} & A plan that achieves the goal state with the minimum possible cost (such as time or resources).\\
    \midrule
    \textbf{Plan Length} & A numerical value that represents the number of actions or steps required to achieve a given goal.\\
    \midrule
    \textbf{Degree of Correctness} & It is the ratio of solved goals and the total number of goals.\\
    \midrule
    \textbf{Plan Verification Tool} & Determines whether a plan achieves the specified goals while satisfying any constraints and/or requirements.\\
    \bottomrule
  \end{tabular}
  }
\end{table}

Besides being capable of generating natural language content, LLMs have demonstrated remarkable abilities to generate high-quality programming code \citep{vaithilingam2022expectation, tian2023chatgpt} and perform reasoning tasks \citep{huang2022reasoning}. Also, some recent works have used LLMs to guide embodied agents towards user-specified goals using prompting techniques \citep{huang2022language, huang2022inner, ahn2022can, wang2023describe, singh2022progprompt}. These studies evaluate LLMs on the world knowledge already possessed by them and use repeated feedback to help the agents overcome failures and perform actions in common-sense domains. In contrast, automated planning problems require generating a plan by reasoning on the constraints specified in the domain file, which is a different and more challenging task. Recently, some works have also explored the capabilities of LLMs in automated planning. \citet{valmeekam2022large, valmeekam2023planning} proposed a suite of benchmark tests to evaluate the planning capabilities of LLMs. They tested recent models such as GPT-3 \citep{brown2020language}, Instruct-GPT \citep{ouyang2022training}, and BLOOM \citep{scao2022bloom} on these benchmarks and found that they performed poorly, with Instruct-GPT achieving only 5\% valid plans on 500 \texttt{Blocksworld} problems. \citet{silver2022pddl} also prompted GPT-3.5 for plan generation and observed similar results but noted that its performance varied across domains. Other works have used LLMs to generate goals \citep{xie2023translating} or problem files given a domain as input \citep{liu2023llm+} and then used a traditional planner to create plans. 

Despite these and other recent studies exploring the potential of LLMs in automated planning, there is a lack of comprehensive information on the capabilities of LLMs for plan generation. Current approaches only evaluate plan generation capabilities using prompting techniques. To address this gap, this study aims to provide a detailed analysis of the capabilities of LLMs in plan generation and evaluate their generalization capabilities.
\section{Research Questions}

In this study, we aim at exploring the capabilities of LLMs in solving planning problems. To do that, we address the following four research questions (RQ):
\begin{itemize}
    \item \textbf{RQ1 - To what extent can LLMs solve planning problems?} Automated planning requires reasoning abilities to generate a plan, satisfying given constraints. Pretrained LLMs have been shown to reason in recent works, specifically analogical reasoning \citep{agrawal2023llms}, a necessity for automated planning. To answer how well pre-trained (or \textit{vanilla}) LLMs can generate plans, we test state-of-the-art OpenAI models with zero-shot prompting and the others (T5, CodeT5, etc) without fine-tuning. Since most of these models are not familiar with PDDL specifications during training, we expect them to perform poorly without additional fine-tuning for domain adaptation.
We employ a plan verification tool (i.e., VAL \citep{plan-verif-val}) to verify LLM-generated plans and evaluate them on commonly used, planning-related metrics such as satisficing, optimal, invalid plans, and the degree of correctness.
    \item \textbf{RQ2 - What pre-training data is effective for plan generation?} We provide a set of pre-training data from a diverse set of planning domains (with varying difficulty) to answer this question. 
We compare the performance of models pre-trained exclusively on textual corpora with those that incorporate both code and natural language during the pre-training process. Our goal, in addressing this question, is to provide possible directions for future researchers to select the appropriate LLMs for plan generation.
    \item \textbf{RQ3 - Which approach between fine-tuning and prompting improves plan generation?} Our objective is to compare the effectiveness of fine-tuning and prompting approaches for plan generation. Fine-tuning LLM updates the parameters of the model using a labeled dataset from the target task. Prompting controls the input to an LLM using a template or a cue to elicit the desired output. We want to assess whether updating model weights through fine-tuning provides superior domain adaptation compared to prompting LLMs for the desired outcome. 
    \item \textbf{RQ4 - Are LLMs capable of plan generalization?} To the best of our knowledge, the current literature provides a limited understanding of the plan generalization capabilities of LLMs \citep{valmeekam2023planning}. Only a handful of studies have studied and quantitatively measured them. To better evaluate the generalization capabilities of LLMs within the scope of automated planning, we think that the current definition of plan generalization needs to be clarified due to its limited scope as it fails to account for all possible scenarios that may arise when using LLMs to plan. Therefore, we propose 
    three new tasks to quantify the plan generalization capabilities of LLMs accurately. We believe that this new definition can be then used to properly evaluate and categorize the various LLMs approaches.
\end{itemize}
\section{Materials and Methods}

In this section, we describe our planning dataset and discuss the difficulty classification of the planning domains. We also provide an overview of the LLMs being evaluated and the experimental setup for plan generation.
In what follows, let the training dataset be 
$D_{train} = {(x_1, y_1), ..., (x_n, y_n)}$ where each $x_i$ is a planning problem and $y_i$ is the corresponding optimal plan for problem $x_i$. Let the testing dataset be 
$D_{test} = {(x_{n+1}, y_{n+1}), ..., (x_m, y_m)}$ where each $x_i$ is a previously unseen planning problem and $y_i$ is the corresponding optimal plan for problem $x_i$. Note that $D_{train}$ and $D_{test}$ consist of pairs of planning problems and their corresponding optimal plans, generated using FastDownward \citep{helmert2006fast}, a classical planning system based on heuristic search.

\subsection{Planning Dataset}

The International Planning Competition (IPC) \citep{ipc} is a biennial event that evaluates state-of-the-art automated planning and scheduling systems. A new set of planners and planning domains are periodically released as part of the IPC. We consider six classical planning domains represented in PDDL, released as a part of the IPC, to assess planning capabilities in LLMs. We use problem generators that came with these domains \citep{seipp-et-al-zenodo2022} to generate a planning dataset with 18,000 problems for each domain. We use a random seed to generate the problems and further enforced that there are no duplicate problems in the dataset. We generate the ground truth optimal plans for these problems using the planner FastDownward, firstly presented by~\citep{helmert2006fast},  with A\textsuperscript{*} LM-Cut heuristics \citep{helmert2011lm}. Table \ref{difficulty} lists the planning domains considered in this paper and classifies them into three difficulty classes.  
Note that this classification is based on the complexity of the problem in terms of state space and branching factor. Other factors such as observability of the states and concurrency of actions can also be considered but are not in the scope of this paper as we aim to investigate the optimal plan generation capabilities of LLMs.
 Sections 1.3 in the main paper and 2.1 of the supplementary material present additional information on difficulty classification. 


\begin{table}
  \caption{Difficulty of planning domains.}
  \scriptsize
  \label{difficulty}
  \centering
  \begin{tabular}{llp{5cm}p{3cm}}
    \toprule
    Planning Domain & Difficulty & State Space & Branching Factor\\
    \midrule
    \texttt{Ferry} & Easy & $O(2^{n} * 2 * m * n!)$, $n$ is no. of cars, $m$ is no. of locations & $O(n+1)$\\
    \texttt{Blocksworld} & Easy & $O(3^n)$, $n$ is no. of blocks & $O(4n/2 + 1)$\\
    &\\
    \texttt{Miconic} & Medium & $O(n^{(m+1)} * 2^m *m!)$, $n$ is no. of floors, $m$ is no. of passengers & $O(m+1)$\\
    \texttt{Tower of Hanoi} & Medium & $O(3^n)$, $n$ is no. of disks & $O((k-1)k/2)$, $k$ is no. of pegs\\
    \texttt{Grippers} & Hard & $O(2^n * 3^{nr})$, $n$ is no. of balls, $r$ is no. of robots & $O(3nr + r)$\\
    \texttt{Driverlog} & Hard & $O(L^{(D+T+P)} * K^P * D * T * 2^T)$, $L$ is no. of locations, $D$ is no. of drivers, $T$ is no. of trucks, $P$ is no. of packages & $O(L * (D + T + P + DT + TD))$ \\
    \bottomrule
  \end{tabular}
\end{table}

\subsubsection{Data Format for Fine-tuning}

\begin{figure}[t]
  \centering
  \begin{subfigure}[b]{0.45\textwidth}
    \centering
    \includegraphics[width=\textwidth]{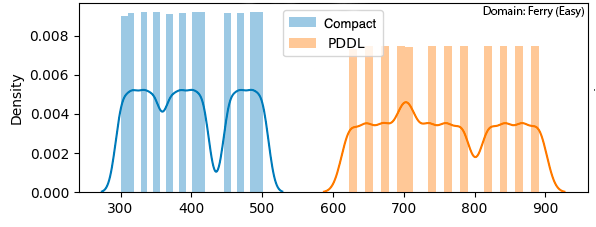}
    \caption{Token lengths for PDDL vs Compact Form}
    \label{fig:pddl-vs-compact-ferry}
  \end{subfigure}
  \begin{subfigure}[b]{0.48\textwidth}
    \centering
    \includegraphics[width=\textwidth]{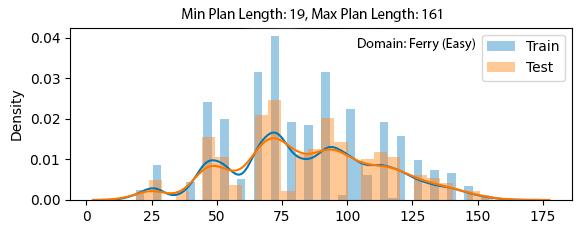}
    \caption{Plan length distribution of train and test sets}
    \label{fig:plantoken-ferry}
  \end{subfigure}
  \caption{(a) Token length for the \texttt{Ferry} domain. Using the compact form achieves an average 40\% reduction in token length in comparison to PDDL. (b) Plan length distribution of both train and test sets are carefully chosen to be similar and within the contextual limitation of the LLMs considered in this paper. Token lengths are in the x-axis and number of problems/plans per token length are in y-axis.}
  \label{fig:ferry-viz}
\end{figure}

Due to the limitation in the amount of contextual information that LLMs take during fine-tuning, it hinders the practical use of the PDDL domain and problem files as input sequences directly. With the domain and problem specifications, it may not be expressive enough to provide the necessary context for the model. To address this limitation, we consider an approach that transforms the PDDL representation into a more compact format with reduced token length without compromising any critical syntactic information. The compact form allows us to maintain the fidelity of the original PDDL representation while making them suitable for fine-tuning LLMs with limitations in the length of the context. Figure \ref{fig:pddl-vs-compact-ferry} illustrates the difference in token lengths between using PDDL and the compact data format for the \texttt{Ferry} domain as an example. We use plans without any modifications because their average token length for the dataset falls within the context length of LLMs as shown in Figure \ref{fig:plantoken-ferry}. To evaluate the plan generation capabilities of the LLMs, we initially use datasets with identical distributions for training and testing. We also report an experiment with out-of-distribution problems. Section 2.2 of the supplementary material contains the details of the compact representation of the PDDL problem and domain files, code, and examples.

\subsubsection{Data Format for Prompting}

Prompting LLMs control the output plan generated using the input sequences. Unlike fine-tuning, prompting allows detailed contextual information as a part of the input sequences. We directly use problem and domain files in PDDL as input. We consider two prompting strategies: zero-shot, and few-shot, to evaluate the planning capabilities in LLMs. In the zero-shot prompting approach, we provide the domain and problem files as input, without any additional examples. In contrast, the few-shot prompting approach provides a few samples from the same domain and its corresponding plan, in addition to the problem and domain files. Section 2.3 of the supplementary material show the example templates used for both prompting techniques. 


\begin{table}
  \caption{LLM architectures used in our study.}
  \scriptsize
  \label{llm-arch}
  \centering
  \begin{tabular}{llllll}
    \toprule
    Model-Parameters & Pre-training & Layers & Heads & Embed. Size & Context Length\\
    \midrule
    T5-base-220M & NL & 12 & 12 & 768 & 512 \\
    CodeT5-base-220M & NL+code & 12 & 12 & 768 & 512 \\
    Codegen-multi-350M & NL+code & 24 & 16 & 1024 & 1024 \\
    text-davinci-03 & NL & NA & NA & NA & 8000 \\
    code-davinci-02 & NL+code & NA & NA & NA & 8000 \\
    \bottomrule
  \end{tabular}
\end{table}

\subsection{Large Language Models: Overview and Considerations}

Table \ref{llm-arch} 
gives an overview of the language models (LLMs) employed in this study, including critical design parameters such as the number of layers, heads, embedding size (head dimension), context length, and data source (natural language (NL) and/or code). Our investigation in this paper involves a combination of LLMs that have been pre-trained on natural language and/or code. Specifically, we utilize two OpenAI models, namely text-davinci-03 and code-davinci-02 for prompting. We set the temperature parameters in these models to 0.2 as increasing the temperature value result in being more hallucinative and prone to taking risks. It is worth noting that both OpenAI models are derived from GPT-3, thereby inheriting its underlying architecture. However, the exact specifications of these models are unknown, and we write it as "not available" (NA) in our table. Our fine-tuning process involves models such as T5, CodeT5, and Codegen trained on the planning dataset (from above) using regularization and weight decay to prevent overfitting. We conduct our experiments using 24 CPU cores, along with a single A100 40GB GPU and 1TB of RAM. On average, it takes less than 2.5 hours to fine-tune a model.


\subsection{Plan Generalization}
\label{sec:generalization}

In this paper, we are interested in the ability of LLMs to generate plans that are close to optimal plans for unseen, out-of-distribution problems from different planning domains. We need a metric to measure the distance between a pair of plans. It is typical to measure the distance between plans based on differences in actions, states, or causal structures \citep{diverse-plan-ijcai2007}. In this paper, we use an action-based criterion for LLM-generated plan vs optimal plan, i.e., the Hamming distance between the order of actions in two plans, motivated by the recent literature \citep{reshaping-diverse-aaai2020}. This metric measures how much  LLMs deviate from the sequences of actions in the optimal plan.
More formally, the plan generalization error $E_{pg}$ is defined as:
{\small
\begin{equation}
E_{pg} = \frac{1}{|m-n|} \sum_{i=n+1}^{m} \frac{H(y_i, \hat{y}_i)}{|A|}
\end{equation}
}
where $\hat{y}_i$ is the generated plan for $x_i \in D_{test}$, $H(\cdot,\cdot)$ is the Hamming distance between two plans, and $|A|$ is the total number of actions in the planning domain. A lower $E_{pg}$ means that LLM can produce plans more similar to the optimal plans generated by a traditional planning system. An LLM with $E_{pg}$ below a certain threshold (in this work, $E_{pg} \leq 0.5$) is considered to have strong plan generalization capabilities. To evaluate the plan generalization capabilities of LLMs and provide insights to address RQ4, we propose the following three tasks:
\begin{itemize}
    \item \textbf{Task 1 - Plan Length Generalization:} We evaluate the ability of LLMs to generalize to plan lengths that are out of the distribution of the training set. Given a set of planning domains in the training set $\mathcal{D}={D_1, D_2, ..., D_n}$, we select a plan length $l_i$ for each domain $D_i$ that is outside the range of plan lengths considered in the training set. We then pose the planning problem for each domain with the selected plan length to the LLM. Let $\hat{\pi}$ denote the plan generated by the LLM, and let $\pi^*$ denote the optimal plan for the planning problem. We evaluate the plan $\hat{\pi}$ on planning-related metrics mentioned in RQ1 (in addition to $E_{pg}$).
    \item \textbf{Task 2 - Object Name Randomization:} We test the LLMs ability to generate plans using randomized object names not present in the training set. We first create a set of randomized object names for each planning domain in the dataset using the IPC problem generator. We then replace the object names in the planning problems in each domain for the test set with the randomized names. We pose the modified planning problems to the LLMs and evaluate the generated plan $\hat{\pi}$ using the set of planning metrics mentioned in Task 1.
    \item \textbf{Task 3 - Unseen Domain Generalization:} We evaluate the LLMs ability to generalize to planning problems from new domains not included in the training set. We select a set of planning domains $\mathcal{D}'={D'_1, D'_2, ..., D'_m}$ that is different from the planning domains in the training set. We pose the planning problems in each domain to the LLM and evaluate the generated plan $\hat{\pi}$ using the planning metrics mentioned in Task 1. 
\end{itemize}

Overall, we believe these tasks provide a comprehensive evaluation of the ability of LLMs to generalize to different types of planning problems.
\section{Experimental Results}

\begin{table}[ht]
  \centering
  \caption{Evaluation of plan generation capabilities of LLMs (both prompting pre-trained model and fine-tuned model). 
  For each model, we report the inference time (Inf. Time), the percentage of satisficing plans (Sat. Plans), the percentage of optimal plans (Opt. Plans), and the degree of correctness (Deg. Corr.).
  }
  \label{llm-eval}
  \resizebox{\textwidth}{!}{%
 . \begin{tabular}{p{2cm}p{1.7cm}p{1.5cm}*{9}{c}}
    \toprule
    \multirow{2}{*}{Models} & \multirow{2}{*}{Type} & \multirow{2}{*}{Inf. Time} & \multicolumn{3}{c}{Sat. Plans (\%)} & \multicolumn{3}{c}{Opt. Plans (\%)} & \multicolumn{3}{c}{Deg. Corr.} \\
    \cmidrule(lr){4-6} \cmidrule(lr){7-9} \cmidrule(lr){10-12}
    & & & E & M & H & E & M & H & E & M & H \\
    \midrule
    \multirow{2}{*}{T5} & Pre-trained & 4.03s & 0 & 0 & 0 & 0 & 0 & 0 & 0 & 0 & 0 \\
     & Fine-tuned & 1.59s & 0.11 & 0 & 1.16 & 0.11 & 0 & 0.36 & 0.02 & 0 & 0.03 \\
     \midrule
    \multirow{2}{*}{CodeT5} & Pre-trained & 4.22s & 2.70 & 0.6 & 0 & 1.73 & 0.6 & 0 & 0.07 & 0 & 0 \\
     & \textbf{Fine-tuned} & \textbf{1.50s} & \textbf{97.57} & \textbf{92.46} & \textbf{89.54} & \textbf{86.21} & \textbf{90.36} & \textbf{66.71} & \textbf{0.99} & \textbf{0.95} & \textbf{0.95} \\
     \midrule
    \multirow{2}{*}{Codegen} & Pre-trained & 4.52s & 1.70 & 0.16 & 0 & 0.17 & 0 & 0 & 0.01 & 0.01 & 0 \\
     & Fine-tuned & 1.93s & 37.95 & 23.74 & 9.83 & 35.33 & 11.67 & 1.92 & 0.67 & 0.32 & 0.07 \\
     \midrule
    \multirow{2}{*}{text-davinci} & Zero-shot & 3.89s & 8.78 & 6.43 & 0 & 3.92 & 0 & 0 & 0.27 & 0.21 & 0 \\
     & Few-shot & 3.88s & 10.82 & 6.90 & 3.31 & 7.23 & 2.84 & 1.21 & 0.32 & 0.17 & 0.04 \\
     \midrule
    \multirow{2}{*}{code-davinci} & Zero-shot & 3.58s & 17.42 & 11.77 & 4.38 & 8.23 & 6.11 & 1.84 & 0.38 & 0.27 & 0.21 \\
     & Few-shot & 3.51s & 23.52 & 17.48 & 11.89 & 17.57 & 8.85 & 4.69 & 0.57 & 0.32 & 0.28 \\
    \bottomrule
  \end{tabular}
  }
\end{table}

In this section, we present the quantitative and qualitative results obtained using LLMs to generate plans for classical planning domains of varying difficulty and generalize to unseen domains. All the reported results in this paper are averaged over five independent runs.
We evaluate the performance of the considered LLMs on a test set with 3600 planning problems per domain classified into easy (\textit{E}), medium (\textit{M}), and hard (\textit{H}) categories. The problems exhibit the same distribution of plan length as the training set but consist of a distinct set of problem instances not encountered during the training phase. We assess the generated plans using various planning-related metrics, including satisficing plans (\textit{Sat. Plans}), optimal plans (\textit{Opt. Plans}), degree of correctness (\textit{Deg. Corr.}), and inference time (\textit{Inf. Time}). We can determine the number of invalid plans by subtracting the percentage of satisficing plans from 100 percent. 

\subsection{Results on Plan Generation}

Table \ref{llm-eval} indicates that code-davinci (zero-shot) performs better among the vanilla models. However, \textbf{we observe limited planning capabilities overall in pre-trained models}, confirming the findings of \citep{valmeekam2022large}.

Table \ref{llm-eval} also shows that the fine-tuned CodeT5 model performs best across all considered domains, with the least inference time. {\em  Overall, it has been observed that fine-tuned language models (LLMs) are capable of generating outputs for planning problems at a rate four times faster than pre-trained LLMs}. This phenomenon can be attributed to the ability of fine-tuned LLMs to discern when to cease output generation, a less developed capability in pre-trained LLMs, as illustrated in Figure \ref{fig:inference_time}. The findings from Table \ref{llm-eval} provide insightful observations that can aid in selecting the most appropriate LLMs for near-optimal plan generation. Our evaluation metrics suggest that \textbf{LLMs pre-trained on programming code outperform those solely trained on the textual corpus}. This addresses our research question RQ2. 
However, although Codegen is a decoder-only model similar to GPT-2, our evaluation indicates that it struggles to generate near-optimal plans after fine-tuning. This is due to Codegen's lack of a dedicated encoder, which may hinder its ability to understand and represent the input prompt. In contrast, CodeT5 leverages both an encoder and a decoder, which allows for better code understanding and generation tasks. Furthermore, CodeT5 incorporates code-specific knowledge by proposing a novel masked identifier prediction task. This task masks all identifiers present in the input prompt. It employs a sentinel token for all occurrences of a specific identifier, allowing the model to better distinguish and recover identifiers in code, which are crucial for code semantics and execution. In this paper, we consider the identifiers for planning domains such as action names, predicates, and object names.

\begin{figure}[ht]
    \scriptsize
    \centering
    \begin{minipage}{0.9\textwidth}
        \colorbox{red!20}{\parbox{\linewidth}{\textbf{Pre-trained CodeT5 (\textit{Inference Time:} 4.12s)} \\
        \small \texttt{b4 b3, put-down b4, unstack b2 b1, unstack x, not clear x, not handempty, ontable, clear, holding x, not x, handempty, ontable x, stack x, not x, not clear x,y, unstack b2 b3, put-down b} \textcolor{red}{\ding{55}}}} \\
        
        \colorbox{blue!20}{\parbox{\linewidth}{\textbf{Fine-tuned CodeT5 (\textit{Inference Time:} 1.01s)} \\
        \small \texttt{unstack b4 b3, put-down b4} \textcolor{green!60!black}{\ding{51}}}}
    \end{minipage}
    \caption{Incoherent generations from pre-trained models leading to longer inference time.}
    \label{fig:inference_time}
\end{figure}

Prompting LLMs such as the OpenAI models have shown to perform better when they have coding-related knowledge. However, they still perform worse than fine-tuning LLMs pre-trained on code. Our observations indicate that \textbf{fine-tuning is a superior approach to solving planning problems with LLMs}. This is attributed to fine-tuning allowing for more customization and adaptation of the model to a specific task, while prompting relies on the general knowledge and competence of the pre-trained models. However, in recent times, prompt engineering has become a widely researched topic, and new techniques are being developed rapidly. Exploring other prompting approaches, such as prompt-tuning \citep{anil2022exploring}, would be interesting. Prompt-tuning learns continuous prompts optimized end-to-end over a training dataset and is as effective as fine-tuning natural language understanding tasks. Although our evaluation addresses RQ3,
we believe there needs to be more extensive research to be done in order to effectively answer this research question. The present study commenced by conducting an initial analysis to explore the potential of LLMs for automated planning. 
In order to address RQ4, we report the results for the the tasks described in Section \ref{sec:generalization}. 

\subsection{Results on Plan Generalization}

\begin{figure}[t]
    \centering
    \includegraphics[width=1\textwidth]{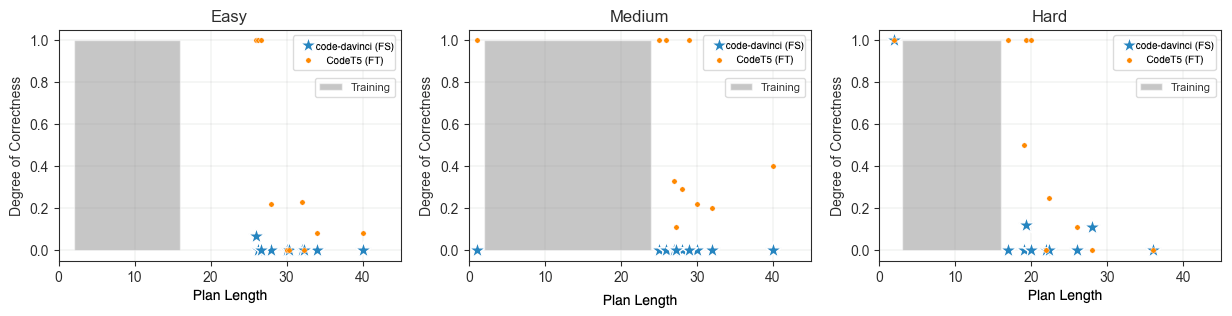}
    \caption{Fine-tuned CodeT5 and code-davinci with few-shot prompting show poor plan length generalization capabilities: $E_{pg}$ for both models is higher than the threshold (0.5), but plans from fine-tuned CodeT5 overall have a higher degree of correctness. The x-axis represents the plan length and the y-axis represents the degree of correctness. The training plan lengths are highlighted in grey.}
    \label{fig:len_gen}
\end{figure}

\noindent \textbf{Task 1 - Plan Length Generalization.}
In the context of generalization experiments, we consider the most proficient models obtained through fine-tuning and prompting methodologies. Specifically, we employ the fine-tuned (FT) CodeT5 and few-shot (FS) prompting of code-davinci approaches. We subject these models to an empirical evaluation by testing them on ten problems per difficulty class. Notably, the ten problems selected are characterized by a plan length outside the distribution of the training set. Figure \ref{fig:len_gen} depicts the outcomes of the plan length generalization assessment across the three different difficulty classes. Our findings demonstrate that the \textbf{fine-tuned CodeT5 model can generalize to plan lengths to some extent, while the few-shot prompting of code-davinci generates only a single valid plan} for a hard domain having a short plan. While neither model produces optimal plans, we observe that the average correctness score of the plans generated by the fine-tuned CodeT5 model is 0.46, while that of code-davinci is 0.04. With regard to $E_{pg}$, the fine-tuned CodeT5 model has an average score of 0.69, whereas code-davinci has an average score of 1. Notably, neither model meets the $E_{pg}$ threshold for generalization. In future work, we aim to investigate recent approaches, such as scratchpad fine-tuning and prompting methodologies \citep{anil2022exploring}, that have been shown to enhance the length generalization capabilities of LLMs.

\begin{table}
  \centering
  \caption{Evaluating the capabilities of LLMs in handling randomized object names.}
  \label{obj_rand}
\resizebox{\textwidth}{!}{%
 \begin{tabular}{cccccccccccc}
    \toprule
    \multirow{2}{*}{Object Names} & \multirow{2}{*}{Model} & \multirow{2}{*}{ST} & \multicolumn{3}{c}{Sat. Plans (\%) / Opt. Plan (\%)} & \multicolumn{3}{c}{Deg. Corr.} & \multicolumn{3}{c}{$E_{pg}$} \\
    \cmidrule(lr){4-6} \cmidrule(lr){7-9} \cmidrule(lr){10-12}
    & & & E & M & H & E & M & H & E & M & H \\
    \midrule
    \multirow{4}{*}{Version 1} & \multirow{2}{*}{CodeT5(FT)} & \textcolor{red}{\ding{55}} & 47.12\% / 33.78\% & 42.74\% / 38.68\% & 39.58\% / 21.22\% & 0.57 & 0.51 & 0.51 & 0.82 & 0.84 & 0.84 \\
     & & \textcolor{green!60!black}{\ding{51}} & 97.52\% / 86.21\% & 92.46\% / 90.36\% & 89.12\% / 66.71\% & 0.98 & 0.95 & 0.95 & 0.15 & 0.18 & 0.18 \\
     \cmidrule(lr){2-12}
    & \multirow{2}{*}{code-davinci(FS)}& \textcolor{red}{\ding{55}} & 23.52\% / 17.57\% & 17.48\% / 8.85\% & 11.89\% / 4.69\% & 0.57 & 0.32 & 0.28 & 0.77 & 0.83 & 0.96 \\
 & & \textcolor{green!60!black}{\ding{51}} & 23.52\%/ 17.57\%	& 17.48\% / 8.85\% & 11.89\% / 4.69\% & 0.57 & 0.32 & 0.28 & 0.77 & 0.83 & 0.96\\
 \midrule
     \multirow{4}{*}{Version 2} & \multirow{2}{*}{CodeT5(FT)} & \textcolor{red}{\ding{55}} & 66.01\% / 64.72\% & 61.98\% / 52.80\% & 55.17\% / 37.5\% & 0.79 & 0.72 & 0.58 & 0.47 & 0.49 & 0.67 \\
     & & \textcolor{green!60!black}{\ding{51}} & 97.52\%/86.21\% & 92.46\%/90.36\% & 89.12\%/66.71\% & 0.98 & 0.95 & 0.95 & 0.15 & 0.18 & 0.18 \\
     \cmidrule(lr){2-12}
    & \multirow{2}{*}{code-davinci(FS)}& \textcolor{red}{\ding{55}} & 23.52\%/17.57\% & 17.48\%/8.85\% & 11.89\%/4.69\% & 0.57 & 0.32 & 0.28 & 0.77 & 0.83 & 0.96 \\
 & & \textcolor{green!60!black}{\ding{51}} & 23.52\%/17.57\%	& 17.48\%/8.85\% & 11.89\%/4.69\% & 0.57 & 0.32 & 0.28 & 0.77 & 0.83 & 0.96\\
  \midrule
     \multirow{4}{*}{Version 3} & \multirow{2}{*}{CodeT5(FT)} & \textcolor{red}{\ding{55}} & 11.82\% / 2.10\% & 4.92\% / 1.47\% & 0.17\% / 0\% & 0.24 & 0.04 & 0.01 & 0.87 & 0.95 & 1 \\
     & & \textcolor{green!60!black}{\ding{51}} & 97.52\%/86.21\% & 92.46\%/90.36\% & 89.12\%/66.71\% & 0.98 & 0.95 & 0.95 & 0.15 & 0.18 & 0.18 \\
     \cmidrule(lr){2-12}
    & \multirow{2}{*}{code-davinci(FS)}& \textcolor{red}{\ding{55}} & 23.52\%/17.57\% & 17.48\%/8.85\% & 11.89\%/4.69\% & 0.57 & 0.32 & 0.28 & 0.77 & 0.83 & 0.96 \\
 & & \textcolor{green!60!black}{\ding{51}} & 23.52\%/17.57\%	& 17.48\%/8.85\% & 11.89\%/4.69\% & 0.57 & 0.32 & 0.28 & 0.77 & 0.83 & 0.96\\
 \bottomrule
  \end{tabular}
}
\end{table}

\noindent \textbf{Task 2 - Object Name Randomization.}
For object name randomization, we created three versions of randomized variables to evaluate the plan generalization capabilities of  LLMs. In version 1, we used only single-digit numeric values as object names. In version 2, we used alphanumeric strings of length 2 (similar to the convention followed by IPC generators), where the combinations of alphabets and numerals used were unseen during training. Version 3 consisted of object names named after three alphabets. For the few-shot prompting of code-davinci, we used an example from the training set and posed a problem with randomized object names for which the plan needed to be generated. We also implemented a symbol table that maps the randomized object names to the same vocabulary as the training set to comprehensively evaluate the dependence of LLMs on the training data or the prompting example for plan generation. Table \ref{obj_rand} captures the performance of the models considered for generalization, and it is observed that code-davinci's performance is unaffected by object name randomization and retains the same performance as seen in Table \ref{llm-eval}. However, code-davinci has a high $E_{pg}$, showing poor plan generalization capabilities. Fine-tuned CodeT5 has the best performance for version 2 and better plan generalization capabilities than any other model (without a symbol table). We observed a decrease in performance when the length of object names was increased, as in version 3, because the model confuses action names and object names. We further noticed an improvement in plan generalization capabilities when a symbol table (\textit{ST}) was used to map the problem name with randomized object names to the same vocabulary as the training set.

\begin{figure}[h]
    \scriptsize
    \centering
    \begin{minipage}{0.95\textwidth}
        \colorbox{green!15}{\parbox{\linewidth}{\textbf{Ground Truth Plan - $\pi^*$ (\textit{Domain:} \texttt{logistics})} \\
        \small \texttt{load-airplane p2 a0 l1-0, fly-airplane a0 l1-0 l0-0, unload-airplane p2 a0 l0-0, load-airplane p0 a0 l0-0, fly-airplane a0 l0-0 l1-0, unload-airplane p0 a0 l1-0}}} \\
        
        \colorbox{red!13}{\parbox{\linewidth}{\textbf{Fine-tuned CodeT5 (\textit{Inference Time:} 1.12s)} \\
        \small \texttt{load-airplane driver0 l0, \textcolor{red}{load-truck package3 truck0 s0}, fly-airplane truck0 s0 s1 driver0, \textcolor{red}{load-truck package2 truck0 s1}, unload-airplane p0 a0 l1-0, \textcolor{red}{drive-truck truck0 s1 s0 driver0}, \textcolor{red}{unload-truck package2 truck0 s0}} \textcolor{red}{\ding{55}}}} \\
        
        \colorbox{red!13}{\parbox{\linewidth}{\textbf{Few-shot prompting code-davinci (\textit{Inference Time:} 3.51s)} \\
        \small \texttt{load-truck p0 t0 l0-0, load-truck p1 t1 l1-0, drive-truck t0 l0-0 l1-0 c1, unload-truck p0 t0 l1-0, drive-truck t1 l1-0 l0-0 c0, unload-truck p1 t1 l0-0, load-airplane p2 a0 l1-0} \textcolor{red}{\ding{55}}}}
    \end{minipage}
    \caption{Example of an incorrect generations from LLMs for a problem from an unseen domain.}
    \label{fig:unseen_domain_gen}
\end{figure}

\noindent \textbf{Task 3 - Unseen Domain Generalization}
We conducted a study to assess the efficacy of fine-tuned CodeT5 and code-davinci with few-shot prompting in generating plans for domains not included in the training set. Specifically, we selected three new classical planning domains, namely \texttt{childsnack}, \texttt{depots}, and \texttt{satellite}, and created ten problems per domain. We randomly chose an example prompt and its corresponding plan from the six training domains for each problem. We report the results averaged over five random seeds. Our findings reveal that both models failed to generate valid plans in this task, resulting in an $E_{pg}$ of 1. We observed that fine-tuned CodeT5 often confused the action and object names present in the test with those seen during training, showing no capabilities in generalizing to unseen domains. On the other hand, code-davinci generated relevant actions but incorrect plans for all test cases. To further illustrate our observations, we present in Figure \ref{fig:unseen_domain_gen} a comparison between the ground truth plan generated by a planner for the \texttt{logistics} domain and the output produced by the considered LLMs. This comparison highlights the incorrect combination of action and object names. Our experimental results show that fine-tuning LLMs pre-trained on code can significantly improve the generation of near-optimal plans for problems within the training domains. We have also observed that these fine-tuned models exhibit some generalization capabilities, such as handling plans of varying lengths and object names randomized during testing. However, we have found that these models struggle to generate plans for unseen domains. While prompting LLMs like code-davinci show some promising developments in planning capabilities, they currently fall short in generating plans. We hope these experiments will be an essential reference for upcoming researchers to refine their selection of LLMs for planning and explore avenues for improving their plan generation capabilities. Such endeavors will ultimately enhance the potential for complex problem-solving and reasoning capabilities.

\section{Conclusion and Future Work}

Our study in this paper explores the potential of LLMs in solving automated planning problems. To do this, we defined four research questions and addressed them through a comprehensive experimental analysis of several LLMs and a diverse set of planning domains.
The study finds that: 
\begin{enumerate*}[label=\textbf{(\arabic*)}]
    \item Off-the-shelf, pre-trained LLMs are not capable of effectively solving planning problems.
    \item LLMs pre-trained on both natural language and programming code are more capable of plan generation than natural language-only models.
    \item Fine-tuning contributes to improved plan generation.
    \item LLMs have limited plan generalization abilities.
\end{enumerate*}
Moreover, our research highlights that fine-tuning aids in partial generalization to plan lengths not encountered during the training phase while maintaining a higher level of correctness than prompting. Notably, when object names are randomized, fine-tuned models exhibit satisfactory performance only when the randomized vocabulary aligns with the training set. Both prompting and fine-tuning approaches prove ineffective when solving problems from unfamiliar domains. In future work, we plan to investigate recent techniques, such as scratchpad fine-tuning and prompting methodologies, that have been shown to enhance the length generalization capabilities of LLMs. These methods could improve the planning capabilities of LLMs and open up new avenues for their use in solving complex planning problems.



\bibliographystyle{plainnat}
\bibliography{references.bib}

\end{document}


\section*{Supplementary Material}

\tableofcontents
\section{Frequently Asked Questions}

\subsection{What are the main contributions of this paper?}
Our main contributions in this paper are as follows:
\begin{itemize}
    \item  Release a dataset of benchmark classical planning problems for six domains and a publicly available codebase with fine-tuned model weights for future research on LLMs for planning.
    \item We define plan generalization and design new tasks to measure it.
    \item We perform a thorough empirical analysis of LLMs on planning-related metrics and provide answers for four questions - \textbf{(1)} To what extent can LLMs solve planning problems?, \textbf{(2)} What pre-training data is effective for plan generation?, \textbf{(3)} Which approach between fine-tuning and prompting improves plan generation?, and \textbf{(4)} Are LLMs capable of plan generalization?
\end{itemize}

\subsection{Will the codebase be made publicly available?}

Yes, the codebase along with all the fine-tuning, prompting, and inference scripts along with the fine-tuned model weights for T5, CodeT5, and Codegen will be released for public access after the review period in accordance with the  Findability, Accessibility, Interoperability, and Reuse (FAIR) of digital assets principles. The codebase will be made available via Github and the datasets as well as model weights will be released on Zenodo.

\subsection{What is the reason for difficulty classification of the chosen six domains?}
The classification of domain difficulty is contingent upon various factors, including the dataset utilized for the six domains, as well as the branching factor and state space, as detailed in Table 2 of the main paper. Additionally, the difficulty classification takes into account the number of generated and evaluated states using the A\textsuperscript{*} + LM-Cut heuristic in conjunction with the FastDownward planner. Generated states refer to all the possible states that can be reached from the initial state of a planning problem by applying valid actions. These generated states may include redundant or irrelevant states that do not contribute to finding a solution to the problem. Evaluated states, on the other hand, refer to the subset of generated states that have been examined by the search algorithm in order to determine their suitability for inclusion in the final solution plan. These evaluated states are typically evaluated using a heuristic function that estimates the distance from a given state to the goal state. The A\textsuperscript{*} + LM-Cut heuristic is one such function that can be used to efficiently evaluate states and guide the search towards finding an optimal solution. To enhance the justification for the chosen difficulty classification, Table \ref{difficulty-extension} displays the average number of generated and evaluated states for each of the six domains. The data was collected from a total of 18000 problems per domain, arranged in ascending order.

\begin{table}
  \caption{Difficulty classification of planning domains considering the average generated and evaluated states using A\textsuperscript{*}+LM-Cut with FastDownward planner}
  \label{difficulty-extension}
  \centering
  \begin{tabular}{llll}
    \toprule
    Planning Domain & Generated States & Evaluated States & Difficulty\\
    \midrule
    \texttt{Ferry} & 47 & 21 & Easy\\
    \texttt{Blocksworld} & 51 & 35 & Easy\\
    \texttt{Tower of Hanoi} & 141 & 55 & Medium\\
    \texttt{Miconic} & 197 & 72 & Medium\\
    \texttt{Driverlog} & 707 & 334 & Hard\\
    \texttt{Grippers} & 33520 & 1347 & Hard\\
    \bottomrule
  \end{tabular}
\end{table}

\begin{table}[b]
  \centering
  \caption{Variations in the performance of satisficing plans generated by fine-tuned CodeT5 using different dataset sizes per domain. After reaching 18,000 datapoints per domain, i.e., a total of 108,000 total problems belonging to six domains, we see a saturation in the performance of plan generation with increase in data points. The values in the brackets show the increase/decrease in performance in comparison with the values in bold to showcase fixed point iteration.}
  \label{18000}
\begin{tabular}{ccccc}
\toprule
    \multirow{2}{*}{Model} & \multirow{2}{*}{\thead{Dataset size\\(per domain)}}  & \multicolumn{3}{c}{Satisficing Plans (\%)} \\
    \cmidrule(lr){3-5}
    & & Easy & Medium & Hard \\
    \midrule
    \multirow{8}{*}{Fine-tuned CodeT5} & 3000 & 0.20\% & 0\% & 0\% \\
    & 6000 & 19.72\% & 12.11\% & 6.57\% \\
    & 9000 & 47.78\% & 32.65\% & 27.92\% \\
    & 12000 & 69.82\% & 41.17\% & 32.67\%\\
    & 15000 & 86.91\% & 73.88\% & 68.71\% \\
    & \textbf{18000} & \textbf{97.57\%} & \textbf{92.46\%} & \textbf{89.54\%} \\
    & 21000 & 97.57\% \textcolor{green!70!black}{($\uparrow$ 0\%)} & 92.46\% \textcolor{green!70!black}{($\uparrow$ 0\%)} & 89.51\% \textcolor{red}{($\downarrow$ 0.03\%)}\\
    & 24000 & 97.77\% \textcolor{green!70!black}{($\uparrow$ 0.02\%)}& 92.46\% \textcolor{green!70!black}{($\uparrow$ 0\%)}& 89.32\% \textcolor{red}{($\downarrow$ 0.19\%)}\\
    \bottomrule
   \end{tabular}
\end{table}

\subsection{What is the reason for taking 18,000 problems per domain for fine-tuning?}
In our experiment, we examined different numbers of planning problems, starting from 3000 per planning domain and increasing them by multiples of 3. We employed an 80\%-20\% train-test split to evaluate the models. Upon reaching 18,000 problems, we observed the best performance across all the models. Subsequently, we conducted fixed point iteration to assess model performance up to 24,000 problems. However, no significant changes in model performance were observed beyond 18,000 problems. Therefore, we chose 18,000 problems per domain as the de facto standard. In Table \ref{18000}, we show the variation in satisficing plans generated by fine-tuned CodeT5 with respect to the dataset size. We only show the results of fine-tuned CodeT5 for brevity, but similar results are observed across all models considered for fine-tuning.

\subsection{What is the significance of using FastDownward planner to generate the dataset?}

FastDownward is a traditional planning system that searches the space of world states associated with a planning task in the forward direction using heuristics. In the 4th International Planning Competition at ICAPS 2004, FastDownward secured first place in the "traditional (i.e. propositional, non-optimizing) track". FastDownward comes equipped with a variety of search algorithms by default. We utilize the A* + LM-Cut heuristic since it can produce the optimal plans. Thus, we use FastDownward to generate a planning dataset consisting of optimal plans.

\section{Planning Dataset}

This section provides a brief overview of the domains used in our study, accompanied by visualizations of their corresponding PDDL domain and problem files. In addition, we present a compact form for fine-tuning the language models, as well as examples of zero-shot and few-shot prompting.

\subsection{Description of the domains}

\subsubsection{\texttt{Ferry}}

\texttt{Ferry} is a classical planning domain involving a ferry crossing a river to transport passengers and their vehicles from one side of the river. The ferry can carry a limited number of vehicles at a time and must return to the starting point to pick up more cars if passengers are still waiting. The domain includes constraints such as vehicle capacity, safety conditions, and scheduling. The goal of the domain is to transport all passengers and their vehicles to the other side of the river safely and efficiently. The \texttt{Ferry} domain is a benchmark problem for classical planners and has been used in various planning competitions. Based on the domain definition of \texttt{Ferry}, the state space and branching factor can be calculated as follows:

\begin{itemize}
    \item The state space is $O(2^n * 2 * m * n!)$, where $n$ is the number of cars and $m$ is the number of locations. This is because each car can be either on the ferry or at a location ($2^n$ possibilities), the ferry can be at any location ($m$ possibilities), the order of the cars on the ferry matters ($n!$ possibilities), and the ferry can be either empty or full (2 possibilities).
    \item The branching factor is $O(n+1)$, where $n$ is the number of cars. This is because there are only two operators (board and debark) with one parameter each (car) and one operator (sail) with no parameters. For each car, there are two possible operators that can be applied to it (board or debark), but only one of them is applicable at any given state. Additionally, there is one operator that can be applied to any state (sail), which moves the ferry from one location to another.
\end{itemize}

Figure \ref{fig:ferry-pddl} shows the domain and a sample problem file in PDDL for \texttt{Ferry}.

\begin{figure}[t!]
\centering
\small
\begin{minipage}[t]{\textwidth}
\begin{lstlisting}[frame=single,caption={Domain description of \texttt{Ferry}},label=lst:ferry-domain]
(define (domain ferry)
   (:predicates (not-eq ?x ?y)
		(car ?c)
		(location ?l)
		(at-ferry ?l)
		(at ?c ?l)
		(empty-ferry)
		(on ?c))

   (:action sail
       :parameters  (?from ?to)
       :precondition (and (not-eq ?from ?to) 
                          (location ?from) (location ?to) (at-ferry ?from))
       :effect (and  (at-ferry ?to)
		     (not (at-ferry ?from))))


   (:action board
       :parameters (?car ?loc)
       :precondition  (and  (car ?car) (location ?loc)
			    (at ?car ?loc) (at-ferry ?loc) (empty-ferry))
       :effect (and (on ?car)
		    (not (at ?car ?loc)) 
		    (not (empty-ferry))))

   (:action debark
       :parameters  (?car  ?loc)
       :precondition  (and  (car ?car) (location ?loc)
			    (on ?car) (at-ferry ?loc))
       :effect (and (at ?car ?loc)
		    (empty-ferry)
		    (not (on ?car)))))  
\end{lstlisting}
\end{minipage}

\begin{minipage}[t]{\textwidth}
\begin{lstlisting}[frame=single,caption={Problem description of \texttt{Ferry}},label=lst:ferry-problem]
(define (problem ferry-1)
    (:domain ferry)
    (:objects
        location-1 location-2 location-3 car-1 car-2)
    (:init
        (location location-1)
        (location location-2)
        (location location-3)
        (car car-1)
        (car car-2)
        (at car-1 location-1)
        (at car-2 location-1)
        (at-ferry location-2)
        (empty-ferry)
        (not-eq location-1 location-2)
        (not-eq location-1 location-3)
        (not-eq location-2 location-3))
    (:goal
        (and (at car-1 location-2)
             (at car-2 location-2))))
\end{lstlisting}
\end{minipage}

\caption{Domain and problem descriptions of \texttt{Ferry}}
\label{fig:ferry-pddl}
\end{figure}

\begin{figure}[t!]
\centering
\small
\begin{minipage}[t]{\textwidth}
\begin{lstlisting}[frame=single,caption={Domain description of \texttt{Blocksworld}},label=lst:blocksworld-domain]
(define (domain blocksworld)
(:requirements :strips)
(:predicates
(on ?x ?y)
(ontable ?x)
(clear ?x)
(handempty)
(holding ?x))

(:action pick-up
:parameters (?x)
:precondition (and (clear ?x) (ontable ?x) (handempty))
:effect (and (not (ontable ?x))
(not (clear ?x))
(not (handempty))
(holding ?x)))

(:action put-down
:parameters (?x)
:precondition (holding ?x)
:effect (and (not (holding ?x))
(clear ?x)
(handempty)
(ontable ?x)))

(:action stack
:parameters (?x ?y)
:precondition (and (holding ?x) (clear ?y))
:effect (and (not (holding ?x))
(not (clear ?y))
(clear ?x)
(handempty)
(on ?x ?y)))

(:action unstack
:parameters (?x ?y)
:precondition (and (on ?x ?y) (clear ?x) (handempty))
:effect (and (holding ?x)
(clear ?y)
(not (clear ?x))
(not (handempty))
(not (on ?x ?y)))))
\end{lstlisting}
\end{minipage}

\begin{minipage}[t]{\textwidth}
\begin{lstlisting}[frame=single,caption={Problem description of \texttt{Blocksworld}},label=lst:blocksworld-problem]
(define (problem blocksworld-1)
(:domain blocksworld)
(:objects a b c d e f)
(:init
(clear a) (clear b) (clear c)
(ontable d) (ontable e) (ontable f)
(handempty)
(on a b) (on b c)
(on d e) (on e f)
)
(:goal (and (on a c) (on d b) (on e f)))))
\end{lstlisting}
\end{minipage}
\caption{Domain and problem descriptions of \texttt{Blocksworld}}
\label{fig:blocksworld-pddl}
\end{figure}

\subsubsection{\texttt{Blocksworld}}

\texttt{Blocksworld} is a classical planning domain that involves a world of blocks arranged on a table. The domain includes various actions such as picking up, putting down, and stacking or unstacking blocks. The goal of the domain is to reach a specific arrangement of blocks on the table from an initial configuration, which may involve stacking blocks on top of each other or moving blocks to different positions. It is commonly used as a benchmark problem for classical planners and has been used in various planning competitions. The domain is simple and abstract yet complex enough to demonstrate different planning challenges, such as state explosion and search complexity. The state space and branching factor can be calculated as follows:

\begin{itemize}
    \item The state space of the four action \texttt{Blocksworld} is $O(3^n)$ because each block can be in one of three possible locations: on the table, on another block, or in hand. Therefore, the number of possible states is $3^n$, where $n$ is the number of blocks.
    \item The branching factor of the four action \texttt{Blocksworld} is O(4n/2 + 1) because, at each state, four types of actions can be applied to any block: pick up, put down, stack, and unstack. However, not all moves apply to all blocks at all times. For example, a block cannot be picked up if it is unclear, and a block cannot be stacked if the hand is empty. Therefore, the number of applicable actions is at most $4n/2 + 1$, where $n$ is the number of blocks.
\end{itemize}

Figure \ref{fig:blocksworld-pddl} shows the \texttt{Blocksworld} domain and a sample problem file in PDDL. 

\subsubsection{\texttt{Miconic}}
\texttt{Miconic} planning domain is a model of an elevator system that transports passengers between building floors. The elevator can move up or down one floor at a time and can board or depart passengers on each floor. The goal is to deliver all the passengers to their desired floors. The state space and branching factor can be calculated as follows:

\begin{itemize}
    \item The state space of the \texttt{Miconic} domain is $O(n^{(m+1)} * 2^m * m!)$ because each state is determined by the following factors:
    \begin{itemize}
        \item The floor of the elevator ($n$ possibilities, where $n$ is the number of floors)
        \item The destination floor of each passenger ($n$ possibilities for each of the $m$ passengers, where $m$ is the number of passengers)
        \item The location of each passenger (2 possibilities for each passenger: inside or outside the elevator)
        \item The permutation of passengers inside the elevator ($m!$ possibilities, assuming the order matters)
    \end{itemize}
    \item The branching factor of the \texttt{Miconic} domain is $O(m+1)$ because, at each state, there are $m+1$ possible actions: move up, move down, or board/depart a passenger. However, not all actions are applicable at all times. For example, the elevator cannot move up if it is on the top floor and cannot load a passenger if it is full. Therefore, the number of applicable actions is at most $m+1$.
\end{itemize}

Figure \ref{fig:miconic-pddl} shows the \texttt{Miconic} domain and a sample problem file in PDDL.

\begin{figure}[t!]
\centering
\small
\begin{minipage}[t]{\textwidth}
\begin{lstlisting}[frame=single,caption={Domain description of \texttt{Miconic}},label=lst:miconic-domain]
(define (domain miconic)
  (:requirements :strips)
  (:types passenger - object
          floor - object
         )

(:predicates 
(origin ?person - passenger ?floor - floor)
(destin ?person - passenger ?floor - floor)
(above ?floor1 - floor  ?floor2 - floor)
(boarded ?person - passenger)
(not-boarded ?person - passenger)
(served ?person - passenger)
(not-served ?person - passenger)
(lift-at ?floor - floor)
)

(:action board
  :parameters (?f - floor ?p - passenger)
  :precondition (and (lift-at ?f) (origin ?p ?f))
  :effect (boarded ?p))

(:action depart
  :parameters (?f - floor ?p - passenger)
  :precondition (and (lift-at ?f) (destin ?p ?f)
		     (boarded ?p))
  :effect (and (not (boarded ?p))
	       (served ?p)))

(:action up
  :parameters (?f1 - floor ?f2 - floor)
  :precondition (and (lift-at ?f1) (above ?f1 ?f2))
  :effect (and (lift-at ?f2) (not (lift-at ?f1))))

(:action down
  :parameters (?f1 - floor ?f2 - floor)
  :precondition (and (lift-at ?f1) (above ?f2 ?f1))
  :effect (and (lift-at ?f2) (not (lift-at ?f1))))
) 
\end{lstlisting}
\end{minipage}

\begin{minipage}[t]{\textwidth}
\begin{lstlisting}[frame=single,caption={Problem description of \texttt{Miconic}},label=lst:miconic-problem]
(define (problem miconic-1)
(:domain miconic)
(:objects
p1 - passenger p2 - passenger f1 - floor f2 - floor f3 - floor
)
(:init
(lift-at f1)(above f1 f2)(above f2 f3)
(origin p1 f1)(origin p2 f2)(destin p1 f3)
(destin p2 f3)(not-boarded p1)(not-boarded p2)
(not-served p1)(not-served p2)
)
(:goal
(and (served p1) (served p2))
)
)
\end{lstlisting}
\end{minipage}

\caption{Domain and problem descriptions of \texttt{Miconic}}
\label{fig:miconic-pddl}
\end{figure}

\subsubsection{\texttt{Tower of Hanoi}}

The \texttt{Tower of Hanoi} planning domain is a puzzle model involving moving disks of different sizes between pegs. The puzzle starts with all the disks stacked on one peg in decreasing order of size, and the goal is to move all the disks to another peg, following two rules: only one disk can be moved at a time, and a larger disk cannot be placed on top of a smaller disk. The state space and branching factor of this domain can be calculated as follows:

\begin{itemize}
    \item The state space of the \texttt{Tower of Hanoi} is $O(3^n)$ because each disk can be on one of three possible pegs. Therefore, the number of possible states is $3^n$, where $n$ is the number of disks.
    \item The branching factor of the \texttt{Tower of Hanoi} is $O((k-1)k/2)$ because, at each state, there are $k$ possible pegs to move a disk from and $k-1$ possible pegs to move a disk to. Therefore, the number of valid moves is at most $(k-1)k/2$, where $k$ is the number of pegs. However, not all moves are valid, as some may violate the puzzle's rules.
\end{itemize}

Figure \ref{fig:hanoi-pddl} shows the \texttt{Tower of Hanoi} domain and a sample problem file in PDDL. 

\begin{figure}[htbp]
\centering
\small
\begin{minipage}[t]{\textwidth}
\begin{lstlisting}[frame=single,caption={Domain description of \texttt{Tower of Hanoi}},label=lst:hanoi-domain]
(define (domain hanoi)
(:requirements :strips)
(:predicates (clear ?x)
             (on ?x ?y)
             (smaller ?x ?y))

(:action move
:parameters (?disc ?from ?to)
:precondition (and (smaller ?to ?disc) 
                   (on ?disc ?from) 
                   (clear ?disc) 
                   (clear ?to))
:effect  (and (clear ?from) 
              (on ?disc ?to) 
              (not (on ?disc ?from))  
              (not (clear ?to)))
 )) 
\end{lstlisting}
\end{minipage}

\begin{minipage}[t]{\textwidth}
\begin{lstlisting}[frame=single,caption={Problem description of \texttt{Tower of Hanoi}},label=lst:hanoi-problem]
(define (problem hanoi-1)
(:domain hanoi)
(:objects
d1 d2 d3 - disc
p1 p2 p3 - peg
)
(:init
(smaller d1 d2)(smaller d1 d3)(smaller d2 d3)
(on d1 p1)(on d2 p1)(on d3 p1)
(clear p2)(clear p3)(clear d1)
)
(:goal
(and (on d1 p3) (on d2 p3) (on d3 p3))
)
)
\end{lstlisting}
\end{minipage}

\caption{Domain and problem descriptions of \texttt{Tower of Hanoi}}
\label{fig:hanoi-pddl}
\end{figure}

\subsubsection{\texttt{Grippers}}

The \texttt{Grippers} planning domain is a robot model that can move between rooms and pick up or drop balls using its grippers. The robot has two left and right grippers and can hold at most one ball in each gripper. The goal is to move all the balls from one room to another. The state space and branching factor of this domain can be calculated as follows:

\begin{itemize}
    \item The state space of the \texttt{Grippers} domain is $O(2^n * 3^{(nr)})$ because the following factors determine each state:
    \begin{itemize}
        \item The presence of a robot in the room (2 possibilities).
        \item The room of each ball (2 possibilities for each of the $n$ balls, where $n$ is the number of balls)
        \item The gripper of each ball (3 possibilities for each of the $n$ balls: left, right, or none)
        \item The ball in each gripper (n possibilities for each of the $r$ grippers, where $r$ is the number of robots)
    \end{itemize}
    \item The branching factor of the \texttt{Grippers} domain is $O(3nr + r)$ because, at each state, three types of actions can be applied to any ball: pick up, drop, or do nothing. Additionally, there are $r$ possible actions for moving the robot to another room. Therefore, the number of possible actions is at most $3nr + r$.
\end{itemize}

Figure \ref{fig:grippers-pddl} shows the \texttt{Grippers} domain and a sample problem file in PDDL.

\begin{figure}[htbp]
\centering
\small
\begin{minipage}[t]{\textwidth}
\begin{lstlisting}[frame=single,caption={Domain description of \texttt{Grippers}},label=lst:grippers-domain]
(define (domain grippers)
 (:requirements :strips :typing) 
 (:types room object robot gripper)
 (:predicates (at-robby ?r - robot ?x - room)
 	      (at ?o - object ?x - room)
	      (free ?r - robot ?g - gripper)
	      (carry ?r - robot ?o - object ?g - gripper))

   (:action move
       :parameters  (?r - robot ?from ?to - room)
       :precondition (and  (at-robby ?r ?from))
       :effect (and  (at-robby ?r ?to)
		     (not (at-robby ?r ?from))))

   (:action pick
       :parameters (?r - robot ?obj - object ?room - room ?g - gripper)
       :precondition  (and  (at ?obj ?room) (at-robby ?r ?room) (free ?r ?g))
       :effect (and (carry ?r ?obj ?g)
		    (not (at ?obj ?room)) 
		    (not (free ?r ?g))))

   (:action drop
       :parameters (?r - robot ?obj - object ?room - room ?g - gripper)
       :precondition  (and  (carry ?r ?obj ?g) (at-robby ?r ?room))
       :effect (and (at ?obj ?room)
		    (free ?r ?g)
		    (not (carry ?r ?obj ?g)))))
\end{lstlisting}
\end{minipage}

\begin{minipage}[t]{\textwidth}
\begin{lstlisting}[frame=single,caption={Problem description of \texttt{Grippers}},label=lst:grippers-problem]
(define (problem grippers-1)
(:domain grippers)
(:objects
robby - robot
ball cube - object
room-a room-b - room
grip-1 grip-2 - gripper
)
(:init
(at-robby robby room-a)(at ball room-a)(at cube room-b)
(free robby grip-1)(free robby grip-2)
)
(:goal
(and (at ball room-b) (at cube room-a))
)
)
\end{lstlisting}
\end{minipage}

\caption{Domain and problem descriptions of \texttt{Grippers}}
\label{fig:grippers-pddl}
\end{figure}

\begin{figure}[ht]
\centering
\scriptsize
\begin{minipage}[t]{\textwidth}
\begin{lstlisting}[frame=single,caption={Domain description of \texttt{Driverlog}},label=lst:driverlog-domain]
(define (domain driverlog)
  (:requirements :typing) 
  (:types location locatable - object 
          driver truck obj - locatable )
  (:predicates 
		(at ?obj - locatable ?loc - location) (in ?obj1 - obj ?obj - truck)
		(driving ?d - driver ?v - truck) (link ?x ?y - location) 
            (path ?x ?y - location) (empty ?v - truck)
)
(:action load-truck
  :parameters
   (?obj - obj ?truck - truck ?loc - location)
  :precondition
   (and (at ?truck ?loc) (at ?obj ?loc))
  :effect
   (and (not (at ?obj ?loc)) (in ?obj ?truck)))

(:action unload-truck
  :parameters
   (?obj - obj ?truck - truck ?loc - location)
  :precondition
   (and (at ?truck ?loc) (in ?obj ?truck))
  :effect
   (and (not (in ?obj ?truck)) (at ?obj ?loc)))

(:action board-truck
  :parameters
   (?driver - driver ?truck - truck ?loc - location)
  :precondition
   (and (at ?truck ?loc) (at ?driver ?loc) (empty ?truck))
  :effect
   (and (not (at ?driver ?loc)) (driving ?driver ?truck) (not (empty ?truck))))

(:action disembark-truck
  :parameters
   (?driver - driver ?truck - truck ?loc - location)
  :precondition
   (and (at ?truck ?loc) (driving ?driver ?truck))
  :effect
   (and (not (driving ?driver ?truck)) (at ?driver ?loc) (empty ?truck)))

(:action drive-truck
  :parameters
   (?truck - truck ?loc-from - location ?loc-to - location ?driver - driver)
  :precondition
   (and (at ?truck ?loc-from)
   (driving ?driver ?truck) (link ?loc-from ?loc-to))
  :effect
   (and (not (at ?truck ?loc-from)) (at ?truck ?loc-to)))

(:action walk
  :parameters
   (?driver - driver ?loc-from - location ?loc-to - location)
  :precondition
   (and (at ?driver ?loc-from) (path ?loc-from ?loc-to))
  :effect
   (and (not (at ?driver ?loc-from)) (at ?driver ?loc-to)))
)
\end{lstlisting}
\end{minipage}

\begin{minipage}[t]{\textwidth}
\begin{lstlisting}[frame=single,caption={Problem description of \texttt{Driverlog}},label=lst:driverlog-problem]
(define (problem driverlog-1)
  (:domain driverlog)
  (:objects
    driver-1 - driver truck-1 - truck obj-1 - obj loc-1 loc-2 loc-3 - location
  )
  (:init
    (at truck-1 loc-1)(at obj-1 loc-2)(at driver-1 loc-3)(link loc-1 loc-2)(link loc-2 loc-3)(path loc-1 loc-2)(path loc-2 loc-1)(path loc-2 loc-3)(path loc-3 loc-2)(empty truck-1)
  )
  (:goal
    (and
      (at obj-1 loc-1)(at driver-1 loc-1)(empty truck-1)
    )))
\end{lstlisting}
\end{minipage}

\caption{Domain and problem descriptions of \texttt{Driverlog}}
\label{fig:driverlog-pddl}
\end{figure}

\subsubsection{\texttt{Driverlog}}

The \texttt{Driverlog} planning domain is a transportation system model involving drivers, trucks, and packages. The drivers can drive trucks between locations, load and unload packages from trucks, or walk between adjacent locations. The trucks can move between locations if they have a driver. The packages can be loaded or unloaded from trucks at any location. The goal is to deliver all the packages to their destinations. The state space and branching factor of this domain can be calculated as follows:

\begin{itemize}
    \item The state space of the \texttt{Driverlog} domain is $O(L^{(D+T+P)} * K^P * D * T * 2^T)$ because the following factors determine each state:
    \begin{itemize}
        \item The location of each driver, truck, and package ($L$ possibilities for each of the $D+T+P$ entities, where $L$ is the number of locations)
        \item The destination of each package ($K$ possibilities for each of the $P$ packages, where $K$ is the number of possible destinations)
        \item The driver of each truck ($D$ possibilities for each of the $T$ trucks, where $D$ is the number of drivers)
        \item The truck of each driver ($T$ possibilities for each of the $D$ drivers, where $T$ is the number of trucks)
        \item The subset of trucks that are loaded with packages ($2^T$ possibilities, where $T$ is the number of trucks)
    \end{itemize}
    \item The branching factor of the \texttt{Driverlog} domain is $O(L * (D + T + P + DT + TD))$ because, at each state, five types of actions can be applied to any entity: drive, walk, load, unload, or do nothing. However, not all actions are applicable at all times. For example, a driver can only drive a truck at the exact location, and a package can only be loaded if it is clear. Therefore, the number of applicable actions is at most $L * (D + T + P + DT + TD)$, where $L$ is the number of locations.
\end{itemize}

Figure \ref{fig:driverlog-pddl} shows the \texttt{Driverlog} domain and a sample problem file in PDDL.

\subsection{Visualization of compact form}
For fine-tuning LLMs, we make use of the compact form. Figure \ref{fig:compact} shows the compact form representation of the PDDL problems for all the six domains from the training dataset. The implemented python code to perform the conversion of PDDL domain and problem files to compact form is given in Listing \ref{lst:pddl-to-compact}.

\begin{figure}[t]
    \centering
    \begin{minipage}{\textwidth}
        \colorbox{pink!20}{\parbox{\linewidth}{Compact form for \texttt{Ferry} \\
        \scriptsize \texttt{<GOAL> at c0 l0, at c1 l0, at c2 l1 <INIT> location l0, location l1, location l2, location l3, location l4, car c0, car c1, car c2, not-eq l0 l1, not-eq l1 l0, not-eq l0 l2, not-eq l2 l0, not-eq l0 l3, not-eq l3 l0, not-eq l0 l4, not-eq l4 l0, not-eq l1 l2, not-eq l2 l1, not-eq l1 l3, not-eq l3 l1, not-eq l1 l4, not-eq l4 l1, not-eq l2 l3, not-eq l3 l2, not-eq l2 l4, not-eq l4 l2, not-eq l3 l4, not-eq l4 l3, empty-ferry, at c0 l2, at c1 l3, at c2 l3, at-ferry l1 <ACTION> sail <PRE> not-eq from to, location from, location to, at-ferry from <EFFECT> at-ferry to, not at-ferry from <ACTION> board <PRE> car car, location loc, at car loc, at-ferry loc, empty-ferry <EFFECT> on car, not at car loc, not empty-ferry <ACTION> debark <PRE> car car, location loc, on car, at-ferry loc <EFFECT> at car loc, empty-ferry, not on car}}} \\
        
        \colorbox{green!10}{\parbox{\linewidth}{Compact form for \texttt{Blocksworld} \\
        \scriptsize \texttt{<GOAL>on b1 b3, on b2 b4, ontable b3, on b4 b1, on b5 b2, clear b5<INIT>handempty, on b1 b3, on b2 b1, on b3 b5, on b4 b2, clear b4, ontable b5<ACTION> pick-up <PRE> clear x, ontable x, handempty <EFFECT> not ontable x, not clear x, not handempty, holding x <ACTION> put-down <PRE> holding x <EFFECT> not holding x, clear x, handempty, ontable x <ACTION> stack <PRE> holding x, clear y <EFFECT> not holding x, not clear y, clear x, handempty, on x y <ACTION> unstack <PRE> on x y, clear x, handempty <EFFECT> holding x, clear y, not clear x, not handempty, not on x y}}} \\

        \colorbox{blue!10}{\parbox{\linewidth}{Compact form for \texttt{Miconic} \\
        \scriptsize \texttt{<GOAL> served p0, served p1, served p2, served p3, served p4, served p5 <INIT> above f0 f1, above f0 f2, above f0 f3, above f0 f4, above f0 f5, above f0 f6, above f0 f7, above f0 f8, above f0 f9, above f1 f2, above f1 f3, above f1 f4, above f1 f5, above f1 f6, above f1 f7, above f1 f8, above f1 f9, above f2 f3, above f2 f4, above f2 f5, above f2 f6, above f2 f7, above f2 f8, above f2 f9, above f3 f4, above f3 f5, above f3 f6, above f3 f7, above f3 f8, above f3 f9, above f4 f5, above f4 f6, above f4 f7, above f4 f8, above f4 f9, above f5 f6, above f5 f7, above f5 f8, above f5 f9, above f6 f7, above f6 f8, above f6 f9, above f7 f8, above f7 f9, above f8 f9, origin p0 f8, destin p0 f9, origin p1 f0, destin p1 f3, origin p2 f9, destin p2 f8, origin p3 f9, destin p3 f4, origin p4 f6, destin p4 f4, origin p5 f5, destin p5 f0, lift-at f0 <ACTION> board <PRE> lift-at f, origin p f <EFFECT> boarded p <ACTION> depart <PRE> lift-at f, destin p f, boarded p <EFFECT> not boarded p, served p <ACTION> up <PRE> lift-at f1, above f1 f2 <EFFECT> lift-at f2, not lift-at f1 <ACTION> down <PRE> lift-at f1, above f2 f1 <EFFECT> lift-at f2, not lift-at f1}}} \\

        \colorbox{violet!10}{\parbox{\linewidth}{Compact form for \texttt{Tower of Hanoi} \\
        \scriptsize \texttt{<GOAL> on d1 d2, clear d1, on d2 d4, on d3 peg2 , clear d3, on d4 peg1 , on d5 peg3 , clear d5 <INIT> smaller peg1 d1, smaller peg1 d2, smaller peg1 d3, smaller peg1 d4, smaller peg1 d5, smaller peg2 d1, smaller peg2 d2, smaller peg2 d3, smaller peg2 d4, smaller peg2 d5, smaller peg3 d1, smaller peg3 d2, smaller peg3 d3, smaller peg3 d4, smaller peg3 d5, smaller d2 d1, smaller d3 d1, smaller d4 d1, smaller d5 d1, smaller d3 d2, smaller d4 d2, smaller d5 d2, smaller d4 d3, smaller d5 d3, smaller d5 d4, on d1 d2, clear d1, on d2 d5, on d3 peg1 , clear d3, on d4 peg2 , clear d4, on d5 peg3  <ACTION> move <PRE> smaller to disc, on disc from, clear disc, clear to <EFFECT> clear from, on disc to, not on disc from, not clear to}}} \\

        \colorbox{purple!10}{\parbox{\linewidth}{Compact form for \texttt{Grippers} \\
        \scriptsize \texttt{<GOAL> at ball1 room3, at ball2 room2, at ball3 room3, at ball4 room2, at ball5 room3 <INIT> at-robby robot1 room2, free robot1 lgripper1, free robot1 rgripper1, at-robby robot2 room1, free robot2 lgripper2, free robot2 rgripper2, at ball1 room3, at ball2 room1, at ball3 room1, at ball4 room1, at ball5 room3 <ACTION> move <PRE> at-robby r from <EFFECT> at-robby r to, not at-robby r from <ACTION> pick <PRE> at obj room, at-robby r room, free r g <EFFECT> carry r obj g, not at obj room, not free r g <ACTION> drop <PRE> carry r obj g, at-robby r room <EFFECT> at obj room, free r g, not carry r obj g}}} \\

        \colorbox{yellow!10}{\parbox{\linewidth}{Compact form for \texttt{Driverlog} \\
        \scriptsize \texttt{<GOAL> at package1 s2, at package2 s3, at package3 s1, at package4 s3 <INIT> at driver1 s1, at driver2 s2, at driver3 s1, at truck1 s2, empty truck1, at truck2 s2, empty truck2, link s1 s2, link s2 s1, link s1 s3, link s3 s1, at package1 s1, at package2 s1, at package3 s2, at package4 s3 <ACTION> load-truck <PRE> at truck loc, at obj loc <EFFECT> not at obj loc, in obj truck <ACTION> unload-truck <PRE> at truck loc, in obj truck <EFFECT> not in obj truck, at obj loc <ACTION> board-truck <PRE> at truck loc, at driver loc, empty truck <EFFECT> not at driver loc, driving driver truck, not empty truck <ACTION> disembark-truck <PRE> at truck loc, driving driver truck <EFFECT> not driving driver truck, at driver loc, empty truck <ACTION> drive-truck <PRE> at truck loc-from, driving driver truck, link loc-from loc-to <EFFECT> not at truck loc-from, at truck loc-to <ACTION> walk <PRE> at driver loc-from, path loc-from loc-to <EFFECT> not at driver loc-from, at driver loc-to}}} \\
    \end{minipage}
    \caption{Example of compact form representation obtained from PDDL domain and problem files for the six domains considered in the planning dataset. These examples are from the training set.}
    \label{fig:compact}
\end{figure}

\lstset{style=pythonstyle}
\begin{lstlisting}[language=Python, caption={Python code for converting PDDL domain and problem files to compact form},label=lst:pddl-to-compact]
import re
import sys

def find_parens(s):
  """Finds all parentheses in the string `s` and returns a dictionary mapping
  the start index of each parenthesis to its end index.

  Args:
    s: A string.

  Returns:
    A dictionary mapping the start index of each parenthesis to its end index.
  """

  toret = {}
  pstack = []
  flag = 0
  for i, c in enumerate(s):

    if flag == 1 and len(pstack) == 0:
      return toret

    if c == '(':
      pstack.append(i)
      flag = 1
    elif c == ')':
      toret[pstack.pop()] = i

  return toret

def prompt_action(data):
  """Generates a string representation of the action in the given data.

  Args:
    data: A string containing the definition of an action.

  Returns:
    A string representation of the action.
  """

  # Get the name of the action.

  action_name = data.split('\n')[0].split(' ')[1].lower()

  # Get the precondition of the action.

  precondition = data[data.find(':precondition') + 14:data.find(':effect')]
  precondition_parens = find_parens(precondition)
  precondition_strings = []
  for start, end in precondition_parens.items():
    precondition_strings.append(precondition[start:end + 1].strip('()?'))

  # Get the effect of the action.

  effect = data[data.find(':effect') + 10:]
  effect_parens = find_parens(effect)
  effect_strings = []
  for start, end in effect_parens.items():
    effect_strings.append(effect[start:end + 1].strip('()?'))

  # Return a string representation of the action.

  return f'<ACTION> {action_name} {", ".join(precondition_strings)} {", ".join(effect_strings)} </ACTION>'


def prompt_problem(data):
  """Generates a string representation of the problem in the given data.

  Args:
    data: A string containing the definition of a problem.

  Returns:
    A string representation of the problem.
  """

  # Get the initial state of the problem.

  init = data[data.find('(:init') + 8:data.find('(:goal')]
  init_parens = find_parens(init)
  init_strings = []
  for start, end in init_parens.items():
    init_strings.append(init[start:end + 1].strip('()?'))

  # Get the goal state of the problem.

  goal = data[data.find('(:goal') + 7:]
  goal_parens = find_parens(goal)
  goal_strings = []
  for start, end in goal_parens.items():
    goal_strings.append(goal[start:end + 1].strip('()?'))

  # Return a string representation of the problem.

  return f'<INIT> {", ".join(init_strings)} </INIT> <GOAL> {", ".join(goal_strings)} </GOAL>'


def get_prompt(domain_file, problem_file):
  """Generates a string representation of the domain and problem files.

  Args:
    domain_file: The name of the domain file.
    problem_file: The name of the problem file.

  Returns:
    A string representation of the domain and problem files.
  """

  # Read the domain file.

  with open(domain_file, 'r') as f:
    domain_data = f.read()

  # Get the name of the domain.

  domain_name = re.findall(r'(?<=domain )\w+', domain_data)[0]

 

\end{lstlisting}

\subsection{Visualization of prompting techniques}

Figure \ref{fig:zero-shot} shows an example of zero-shot prompting and Figure \ref{fig:few-shot} shows an example of few-shot prompting.

\begin{figure}[t!]
\centering
\scriptsize
\begin{minipage}[t]{1\linewidth}
\begin{lstlisting}[style = normalstyle, frame = single]
The following are the PDDL Domain and Problem files for a classical planning domain - Blocksworld.

PDDL Domain:
(define (domain blocksworld)
  (:requirements :strips)
  (:predicates 
			(on ?x ?y)
			(ontable ?x)
			(clear ?x)
			(handempty)
			(holding ?x)
  )

  (:action pick-up
	     :parameters (?x)
	     :precondition (and (clear ?x) (ontable ?x) (handempty))
	     :effect
	     (and (not (ontable ?x))
		   (not (clear ?x))
		   (not (handempty))
		   (holding ?x)))

  (:action put-down
	     :parameters (?x)
	     :precondition (holding ?x)
	     :effect
	     (and (not (holding ?x))
		   (clear ?x)
		   (handempty)
		   (ontable ?x)))
  (:action stack
	     :parameters (?x ?y)
	     :precondition (and (holding ?x) (clear ?y))
	     :effect
	     (and (not (holding ?x))
		   (not (clear ?y))
		   (clear ?x)
		   (handempty)
		   (on ?x ?y)))
  (:action unstack
	     :parameters (?x ?y)
	     :precondition (and (on ?x ?y) (clear ?x) (handempty))
	     :effect
	     (and (holding ?x)
		   (clear ?y)
		   (not (clear ?x))
		   (not (handempty))
		   (not (on ?x ?y)))))

PDDL Problem:
(define (problem bw-prob1)
       (:domain blocksworld)
       (:objects b1 b2)
       (:init 
       (handempty)
       (ontable b1)
       (clear b1)
       (ontable b2)
       (clear b2)
       )
       (:goal
       (and
       (on b1 b2)
       (clear b1)
       (ontable b2)
       )))
Generate the plan for this PDDL domain and problem:
\end{lstlisting}
\end{minipage}
\caption{An example of zero-shot prompting approach for the domain \texttt{Blocksworld} used with code-davinci and text-davinci LLMs.}
\label{fig:zero-shot}
\end{figure}

\begin{figure}[t!]
\centering
\scriptsize
\begin{minipage}[t]{\textwidth}
\begin{lstlisting}[style = normalstyle, frame = single]
The following is a plan generated for the PDDL Domain and Problem of a classical planning domain - Blocksworld.

PDDL Domain:
(define (domain blocksworld)
  (:requirements :strips)
  (:predicates 
			(on ?x ?y)
                (ontable ?x)
                (clear ?x)
                (handempty)
                (holding ?x)
  )
  (:action pick-up
	     :parameters (?x)
	     :precondition (and (clear ?x) (ontable ?x) (handempty))
	     :effect
	     (and (not (ontable ?x))(not (clear ?x))(not (handempty))(holding ?x)))
  (:action put-down
	     :parameters (?x)
	     :precondition (holding ?x)
	     :effect
	     (and (not (holding ?x))(clear ?x)(handempty)(ontable ?x)))
  (:action stack
	     :parameters (?x ?y)
	     :precondition (and (holding ?x) (clear ?y))
	     :effect
	     (and (not (holding ?x))(not (clear ?y))(clear ?x)(handempty)(on ?x ?y)))
  (:action unstack
	     :parameters (?x ?y)
	     :precondition (and (on ?x ?y) (clear ?x) (handempty))
	     :effect
	     (and (holding ?x)(clear ?y)(not (clear ?x))
		   (not (handempty))(not (on ?x ?y)))))

PDDL Problem:
(define (problem bw-prob1)
       (:domain blocksworld)
       (:objects b1 b2)
       (:init 
       (handempty)(ontable b1)(clear b1)(ontable b2)(clear b2)
       )
       (:goal
       (and(on b1 b2)(clear b1)(ontable b2))))

Plan:
pick-up b1, stack b1 b2

Generate the plan for this new problem from the same domain:

New PDDL Problem:
(define (problem problem_3_1)
(:domain blocksworld)
(:objects b1 b2 b3)
(:init 
    (handempty)
    (ontable b1)
    (ontable b2)
    (clear b2)
    (on b3 b1)
    (clear b3)
)
(:goal
(and
    (ontable b1)(on b2 b1)
    (on b3 b2)(clear b3)
)))

Plan:

\end{lstlisting}
\end{minipage}
\caption{An example of few-shot prompting approach for the domain \texttt{Blocksworld} used with code-davinci and text-davinci LLMs.}
\label{fig:few-shot}
\end{figure}